\documentclass[10pt,twocolumn,letterpaper]{article}

\usepackage{iccv}
\usepackage{times}
\usepackage{epsfig}
\usepackage{graphicx}
\usepackage{amsmath}
\usepackage{amssymb}
\usepackage[numbers]{natbib}
\usepackage[T1]{fontenc}
\usepackage{multirow}
\usepackage{overpic}

\usepackage[breaklinks=true,bookmarks=false]{hyperref}

\iccvfinalcopy % *** Uncomment this line for the final submission

\def\iccvPaperID{****} % *** Enter the ICCV Paper ID here

\ificcvfinal\fi

\begin{document}
%%%%%%%%% TITLE
\title{EISeg: An Efficient Interactive Segmentation Tool based on PaddlePaddle}

% \author{Yuying Hao}
% \author{Yi Liu}
% \author{Yizhou Chen}
% \author{Lin Han}
% \author{Jinrui Ding}

% \affil{haoyuying@baidu.com && liuyi22@baidu.com }

\author{Yuying Hao$^{1}$\quad Yi Liu$^{1}$ \quad Yizhou Chen$^{2*}$
    \quad Lin Han$^{3*}$   \quad Juncai Peng$^{1}$  \quad Shiyu Tang$^{1}$
    \\
    \quad Guowei Chen$^{1}$  \quad Zewu Wu$^{1}$ \quad Zeyu Chen$^{1}$ \quad Baohua Lai$^{1}$ 
    \\
    $^{1}$Baidu, Inc. \quad \quad
    $^2$CQJTU   \quad \quad $^3$NYU\\
    {\tt\small \{haoyuying, liuyi22\}@baidu.com}
}

\maketitle
\ificcvfinal\thispagestyle{empty}\fi
\renewcommand{\thefootnote}{\fnsymbol{footnote}} 
\footnotetext[1]{PaddlePaddle Developers Experts (PPDE)} 

%%%%%%%%% ABSTRACT
\begin{abstract}
In recent years, the rapid development of deep learning has brought great advancements to image and video segmentation methods based on neural networks. However, to unleash the full potential of such models, large numbers of high-quality annotated images are necessary for model training.  Currently, many widely used open-source image segmentation software relies heavily on manual annotation which is tedious and time-consuming. In this work, we introduce EISeg, an Efficient Interactive SEGmentation annotation tool that can drastically improve image segmentation annotation efficiency, generating highly accurate segmentation masks with only a few clicks. We also provide various domain-specific models for remote sensing, medical imaging, 
industrial quality inspections, human segmentation, and temporal aware models for video segmentation. The source code for our algorithm and user interface are available at: \href{https://github.com/PaddlePaddle/PaddleSeg}
{https://github.com/PaddlePaddle/PaddleSeg}.
\end{abstract}

%------------------------------------------------------------------------- 
\section{Introduction}
\label{sec:intro}
As a fundamental task in computer vision, image segmentation is widely used in various fields such as autonomous driving~\cite{self-driving1,self-driving2,liteseg}, medical diagnostics~\cite{medical1}, portrait matting~\cite{matting1, matting2,pp-matting, pp-humanseg}, and so on. In recent years, deep learning has brought tremendous success to computer vision areas. However, it is inherently a data-driven method. A large number of high-quality annotated data is required to train a deep learning model with high accuracy. Datasets for image segmentation 
requires pixel-level precision. Therefore, they are even more complex to annotate compared with other vision tasks. Currently, there are many excellent open-source tools for image segmentation annotation like LabelMe~\cite{labelme}, LabelImage~\cite{labelimage} and CVAT~\cite{cvat}, but they have various limitations. Local software is easy to install and use but relies heavily on manual annotation. Some other open-source tools with advanced features have a large number of dependencies and are difficult to install. These limitations increase the difficulty of using these tools. Adding in more features also complicates the user interface, making the initial learning curve of using such tools steeper. Some online annotation platforms offer a better annotation experience but are not free to use. Uploading data to an online platform can be slow and may cause privacy issues for sensitive data like medical imaging.

\begin{figure}[t!]
  \begin{overpic}[width=\columnwidth]{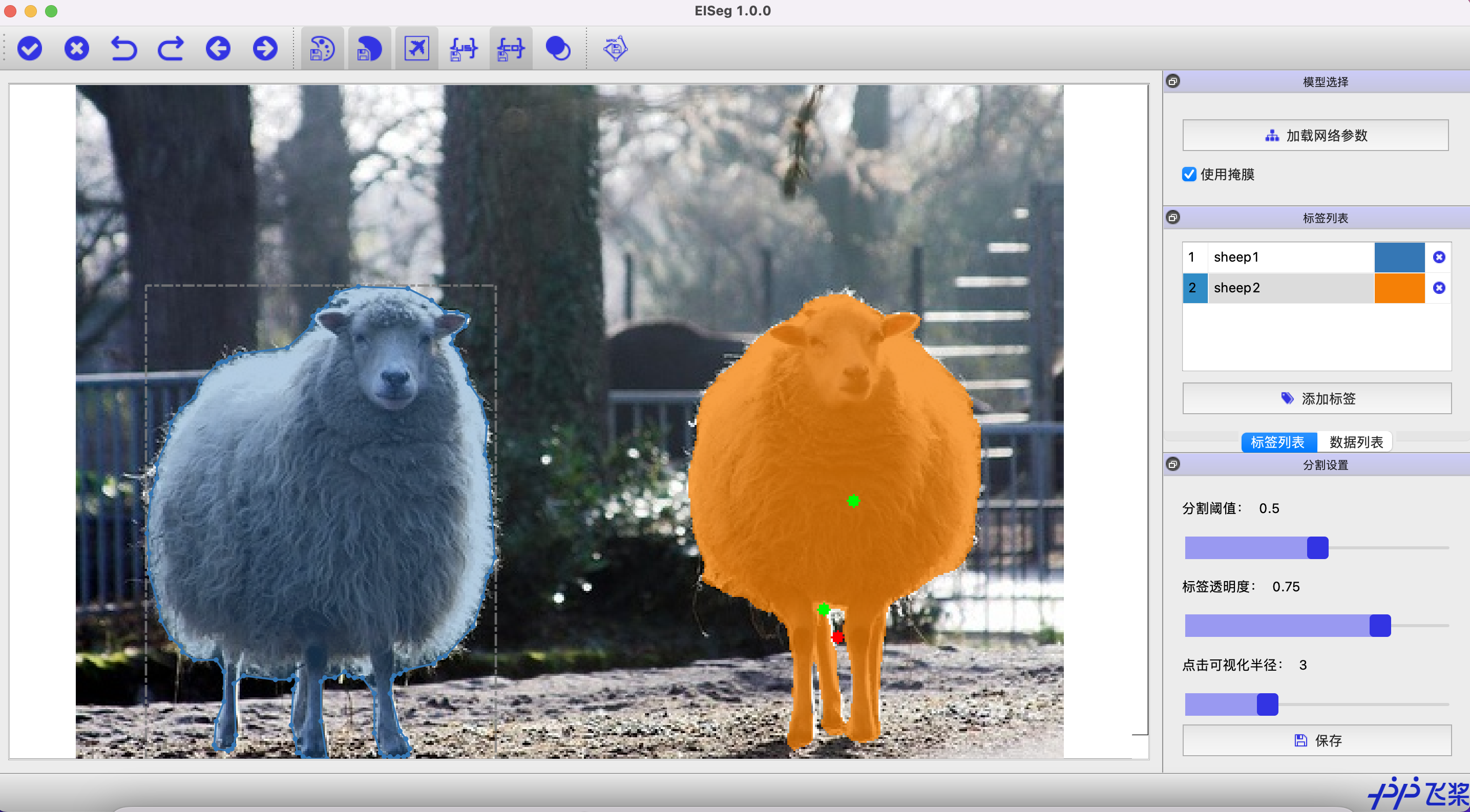}
    \end{overpic}
    \caption{EISeg UI. Users can control the segmentation mask by adding positive (green) and negative (red) clicks. After interactive annotation is finished, press the space key to enter polygon annotation mode. Users can further modify the polygon to tweak the segmentation result.
    }\label{fig:ui}
\end{figure}

To address the issues mentioned above, we develop EISeg. It is easy to install and can boost segmentation efficiency and accuracy drastically with the help of interactive segmentation models. Its user interface is shown in Figure~\ref{fig:ui}. At its core, EISeg is built around interactive segmentation algorithms enabled by Baidu’s deep learning framework PaddlePaddle~\cite{paddlepaddle}. It allows users to quickly extract the object-of-interest with a few clicks. It can drastically improve the efficiency and accuracy of image segmentation annotation compared to manual methods. Shipped as a python package, the software can be installed easily with a single line of command. It supports all major operating systems and can be isolated to a virtual environment to not cause any compatibility issues. The user interface of EISeg is designed to be simple, intuitive, and customizable. Users can change the UI layout and bind keyboard shortcuts to most functionalities in the app to boost productivity. 

\begin{table*}[pth]

\centering
 \caption{EISeg version history and key feature updates} 
 \label{tab:version}
    \begin{tabular}{lll}
    \hline
    \multicolumn{1}{c}{Version} & \multicolumn{1}{c}{Date Released} & \multicolumn{1}{c}{Key Features}                                                                                                                                               \\ \hline
    0.2.0                   & 2021.07.07              & \begin{tabular}[c]{@{}l@{}}(1) First release of interactive segmentation tool EISeg;\\ (2) Released high resolution and lightweight models for general and selfie segmentation.\end{tabular}                                                                                                                                          \\ \hline
    0.3.0                   & 2021.09.16               & \begin{tabular}[c]{@{}l@{}}(1) Added polygon annotation mode, allowing users to adjust interactive model output;\\ (2) Added multi-language user interface;\\ (3) Supported saving segmentation result in grayscale/pseudo color mask and in COCO format.\\ (4) Refined annotation page UI.\end{tabular}                                               \\ \hline
    0.4.0                   & 2021.11.16               & \begin{tabular}[c]{@{}l@{}}(1) Switched to inferencing with static graph model, achieved over 10x speedup;\\ (2) Support remote sensing image segmentation, users can choose bands from RS images;\\ (3) Supported splitting a large image into patches and annotating them one by one;\\ (4) Supported medical imaging, Users can load Dicom files and perform windowlization.\end{tabular} \\ \hline
    0.5.0                   & 2022.04.10               & \begin{tabular}[c]{@{}l@{}}(1) Released models tuned for abdominal organ and spine segmentation;\\ (2) Release models tuned for defect segmentation in aluminum plate visual quality inspection.\end{tabular}                                                                                   \\ \hline
    1.0.0                   & 2022.07.20               & \begin{tabular}[c]{@{}l@{}}(1) Added interactive video segmentation function;\\ (2) Added 3D segmentation function for abdominal organs and spine.\end{tabular}                                                                                       \\ \hline
    \end{tabular}
\end{table*}

Aside from interactive segmentation models for general cases, EISeg also provided several domain-specific models and supported interactive segmentation on videos. More specifically, EISeg provides models specially tuned for remote sensing, industrial quality inspection, and medical imaging datasets. There are also plugin toolboxes implementing features specific to each domain. The video segmentation functionality of EISeg frees users from repeatedly annotating similar objects in nearby frames and can drastically improve video segmentation annotation efficiency. EISeg also implemented support for segmenting 3D medical imaging. We provided models specifically trained for abdominal organ and spine segmentation. 
%The chief focus areas of EISeg included image interactive segmentation, domain-specific interactive segmentation, and video interactive segmentation. We believe this tool will help provide large high-quality image segmentation datasets in various industries. 

The version history and key feature updates of EISeg are shown in Table~\ref{tab:version}. In summary, the features of EISeg include:
\begin{itemize}
\item Improved annotation efficiency and accuracy with the help of interactive segmentation models.
\item Easy to install and use in all major operating systems.
\item Besides models for the general scene, provided domain-specific models in remote sensing, visual quality inspection, 3D medical imaging, and temporal-aware model for videos to accommodate the needs of more industries.
\end{itemize}

\section{Related Works}
Currently, the annotation pipelines of open-source image segmentation annotation tools can be divided into three categories: manual annotation, optimization-algorithm-assisted annotation, and interactive annotation. 

Manual annotation is mainly achieved by encircling the object with a polygon or brushing all the pixels belonging to the object. Representative software in this category includes LabelMe~\cite{labelme}, JingLingBiaoZhu~\cite{jingling}, etc. They are easy to install and intuitive to use. The main drawback is that they rely heavily on manual annotation. The cost of obtaining large high-quality datasets with these softwares are very high.

\begin{figure*}[pth]
	\begin{center}
		%	\fbox{\rule{0pt}{2in} \rule{0.9\linewidth}{0pt}}
		\includegraphics[width=0.9\linewidth]{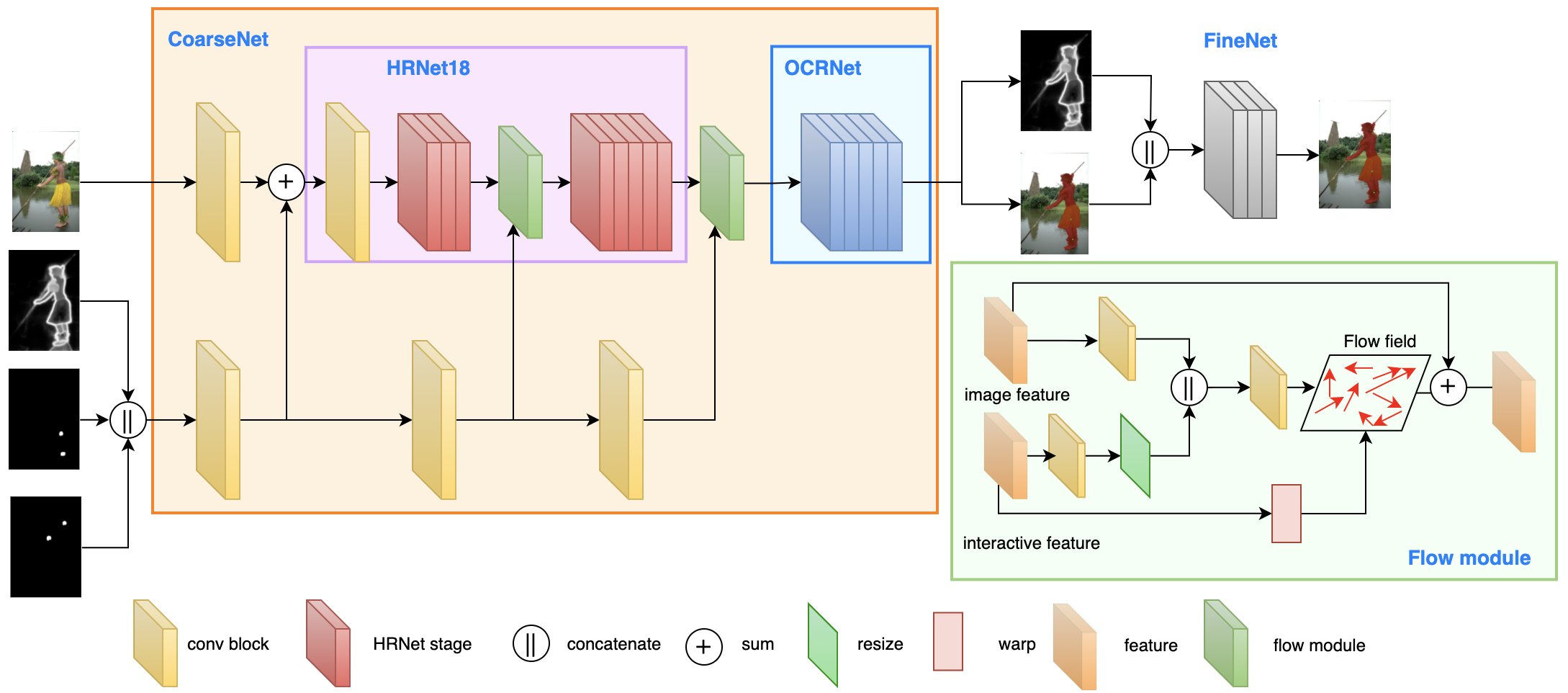}
	\end{center}
	\caption{Overview of EdgeFlow's architecture. We designed a coarse-to-fine network including a CoarseNet and a FineNet. In CoarseNet, we used HRNet-18+OCR as the backbone and appended the edge-guided flow to incorporate interactive information. For FineNet, we used three atrous convolution blocks to refine the coarse segmentation mask. This figure is from the EdgeFlow~\cite{hao2021edgeflow} paper.}
	\label{fig:structure}
	\vspace{-0.15in}
\end{figure*}
Optimization-algorithm-assisted annotation pipelines are mainly based on traditional image processing algorithms like GrabCut~\cite{grabcut} and SLIC~\cite{slic} super-resolution segmentation. Typically, software in this category is also easy to operate. Users would provide scribbles to differentiate foreground from background. The main drawback of these pipelines is that they can only be applied when the image background is relatively uniform and the objects of interest have relatively clear boundaries. The segmentation accuracy of these algorithms is typically not very high. Representative tools in this category include ImageSegmentation~\cite{imagesegmentation} and ImageLabel~\cite{imagelabel}.

Interactive image segmentation pipelines are most often based on deep learning. Based on users’ input, they can generate segmentation masks reliably and flexibly. Users’ input is not in a fixed form. Depending on the algorithm, the input may be points, lines, or bounding boxes. Representative tools in this category include CVAT~\cite{cvat} and this work EISeg. These tools take the image and user's annotations as input and rapidly generate segmentation results. Users are able to guide the models to perform segmentation with positive and negative clicks. There is no longer a need to draw lines or click accurately along objects. The area around a positive or negative control point will be considered foreground or background, respectively. The model produces segmentation results in a semi-automatic fashion based on pre-trained parameters. These pipelines require less user interaction and improve segmentation accuracy, making it more feasible to obtain large high-quality datasets. EISeg provides rich support for industries with non-typical image characteristics or custom annotation pipelines. Besides, EISeg is fast, accurate, and easy to install. Therefore, EISeg is convenient to be applied in image segmentation community. 

\section{Method}

\subsection{Interactive segmentation algorithm}
\label{3.1}

Interactive segmentation algorithms take images and user annotations as input to generate masks of a user interest. This human-in-the-loop method is more flexible compared to non-interactive ones, allowing users to refine segmentation results with additional ROI information. In recent years, interactive segmentation has attracted a lot of attention from professionals from both academia and industry because it can drastically reduce the cost of obtaining large numbers of high-quality segmentation results. In EISeg, the interactive segmentation capability is primarily based on RITM~\cite{ritm} and EdgeFlow~\cite{hao2021edgeflow} which is proposed by Baidu. It has the the following innovations:

\begin{itemize}
\item To make full use of user annotations, EdgeFlow has an early-late-fusion architecture to prevent the fading of user ROI information as it propagates to deeper layers of the network.
\item Instead of using the previous mask from the prior iteration from the model, EdgeFlow used the edge information which tends to be more stable to stabilize model output.
\item In EdgeFlow's annotation pipeline, it’s possible to use both interactive segmentation and polygon-based annotation. This makes the pipeline both efficient and flexible.
\end{itemize}

\begin{figure*}[pth]
	\begin{center}
		\begin{tabular}{cccc}
			
			\includegraphics[width=0.24\linewidth]{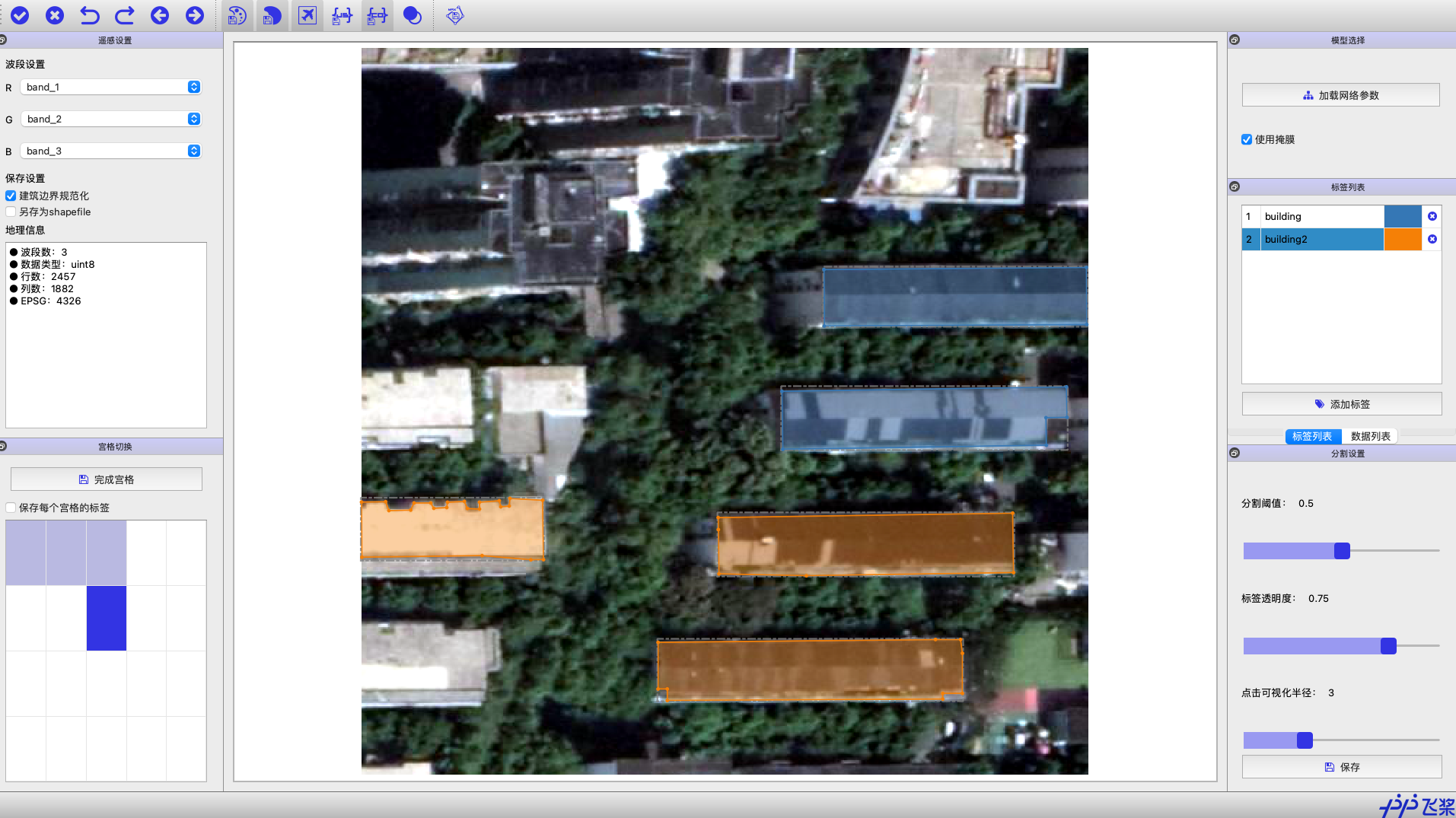}&
			\includegraphics[width=0.24\linewidth]{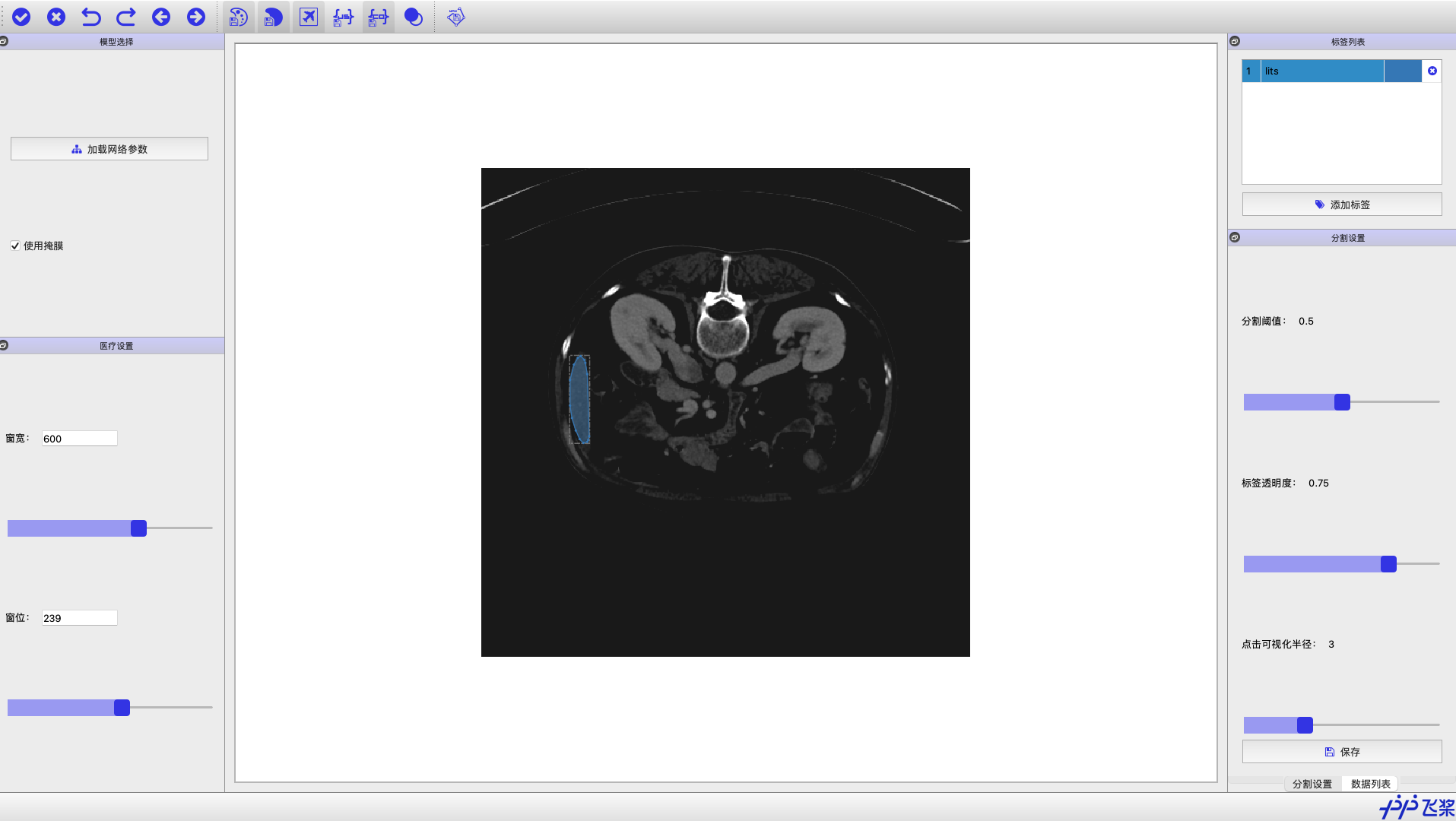}&
			\includegraphics[width=0.24\linewidth]{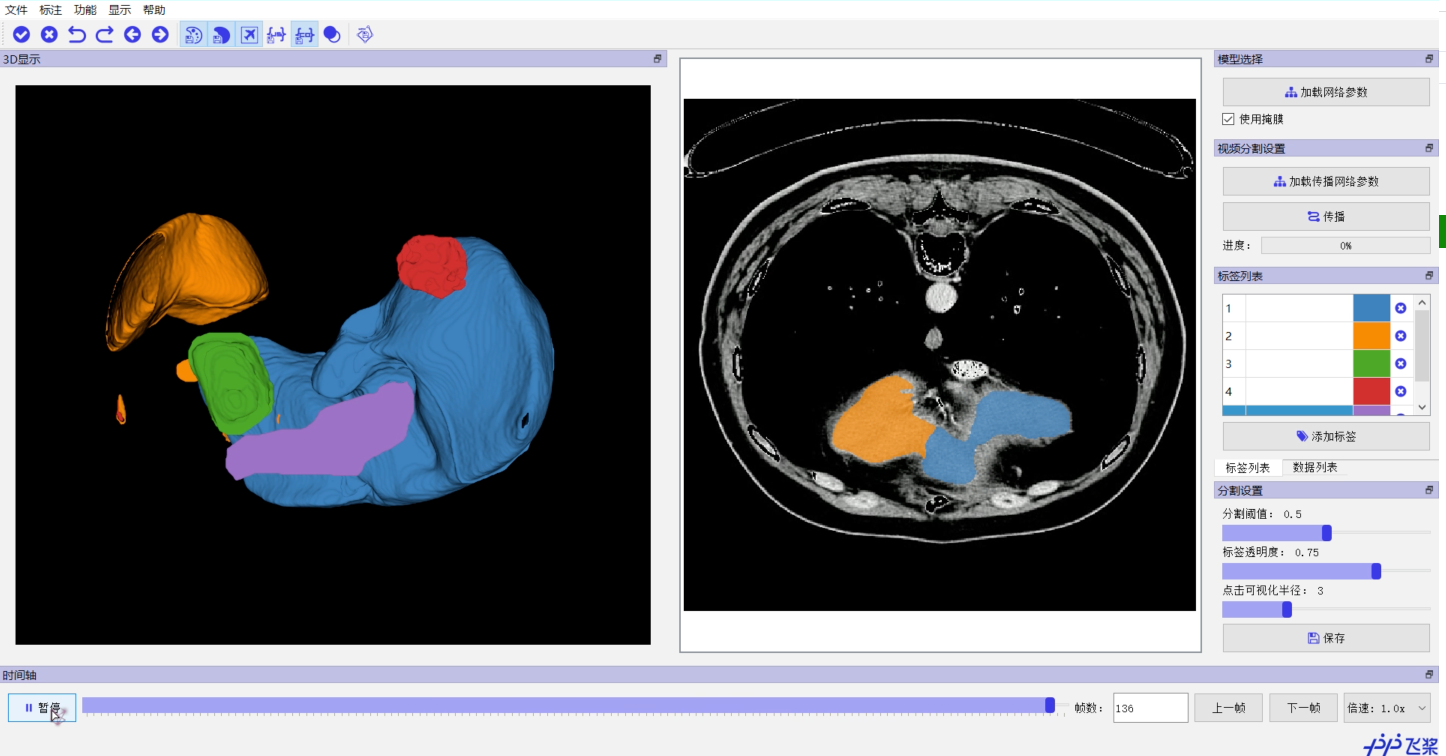}&
			\includegraphics[width=0.24\linewidth]{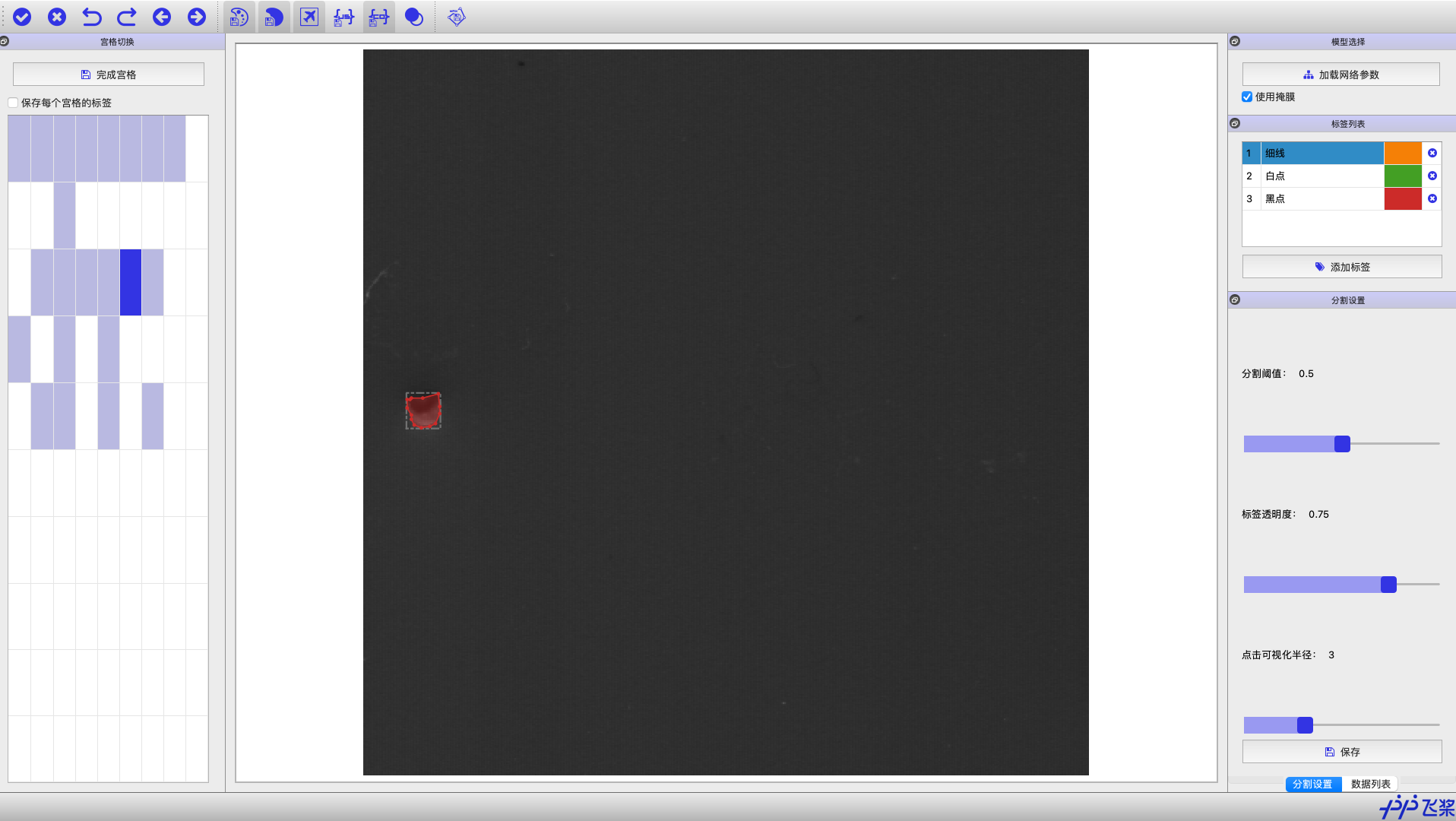}\\
			
		\end{tabular}
	\end{center}
	
    \caption{UI for domain-specific annotation including remote sensing, medical imaging, visual quality inspection, 3D data, etc. For each domain, a toolbox is implemented to better meet the annotation needs.} 
    \label{figshow}
    \vspace{-0.15in}
\end{figure*}
The model architecture of EdgeFlow is shown in Figure~\ref{fig:structure}. An outline of the prediction pipeline is as follows. The model takes an image, the user’s positive and negative clicks, and segmentation edge prior information as input. For the first click, the edge prior input is set to all zeros. In the first stage,  CoarseNet takes the original image, user annotations, and edge prior as input and produces edge information and a coarse segmentation mask. In the second stage, CoarseNet’s output, together with all inputs will be passed to the FineNet to produce a more accurate segmentation result. In each interaction, the edge information from the previous interaction will be inputted into the model as prior to stabilize the model output. Users can interact with the model multiple times until an ideal result is produced.

\subsection{Interactive video segmentation algorithm}
\label{3.2}
The interactive video segmentation algorithm of EISeg has a similar structure to MiVOS~\cite{mivos}. There are three main components in MiVOS which are interactive segmentation, propagation, and fusion component. The general pipeline is as follows. Users would first choose a single frame and perform interactive annotation inside. The frame will be used as a reference frame by the propagation component to propagate segmentation results to nearby frames. Results propagated from different reference frames may disagree inside an intermediate frame. In this case, a fusion model is used to fuse the different masks. This pipeline reduces the need to annotate similar objects repeatedly and can greatly boost video annotation efficiency. 

In the interactive segmentation phase, we reused the image interactive segmentation model in EISeg. It is able to generate accurate reference frames for propagation rapidly and also offers domain-specific models to cover more use cases. In the propagation phase, our method employed a memory bank approach, similar to that in STM~\cite{stm}. We encode history and reference frames and add the encoded features to a memory bank. During propagation, we compared the features of the current frame with those in the memory back and use a decoder to produce segmentation mask for the current frame. The fusion phase resolves conflicts between results propagated from different reference frames. Through experiments, we concluded that a dedicated fusion model outperforms simple approaches like averaging all propagated results. Putting the three phases together, the algorithm is able to achieve state-of-the-art performance.

\subsection{Domain specific models}
\label{3.3}

To meet the annotation need of different industries better, we trained our model on various domain-specific datasets and obtained high-quality results. This currently includes models tuned for drivable free space segmentation in autonomous driving, abdominal organ and spine segmentation in medical imaging, defect segmentation in industrial quality inspection, building segmentation in remote sensing, etc. We also provided features specific to various domains. More concretely

\begin{itemize}
\item For remote sensing annotation, we implemented support for reading and saving images in GTiff and ESRI Shapefile format. For extra large images, we supported dividing the image into small patches before annotation.

\item For 2D medical imaging annotation, we implemented window level and window width adjustment on the images and provided models tuned specifically for abdominal organ and spine segmentation.

\item For 3D medical imaging annotation, we adapted the video annotation pipeline to rapidly perform segmentation on 3D medical imaging. We also implemented 3D visualization of the segmentation results.

\item For defect annotation, we implemented the splitting into patch feature as in remote sensing and supported defect annotation on multiple types of workpieces.

\end{itemize}

\section{Tool usage}

This section details the steps for annotating 2D images, videos and 3D medical images.

\subsection{Image annotation}\label{image_annotation}

The image annotation function supports segmenting common RGB images, grayscale images, remote sensing images in tiff format, medical images in Dicom format, and so on. The user interface is shown in Figure~\ref{fig:ui}. The main steps include:

\begin{figure}[t!]
  \begin{overpic}[width=\columnwidth]{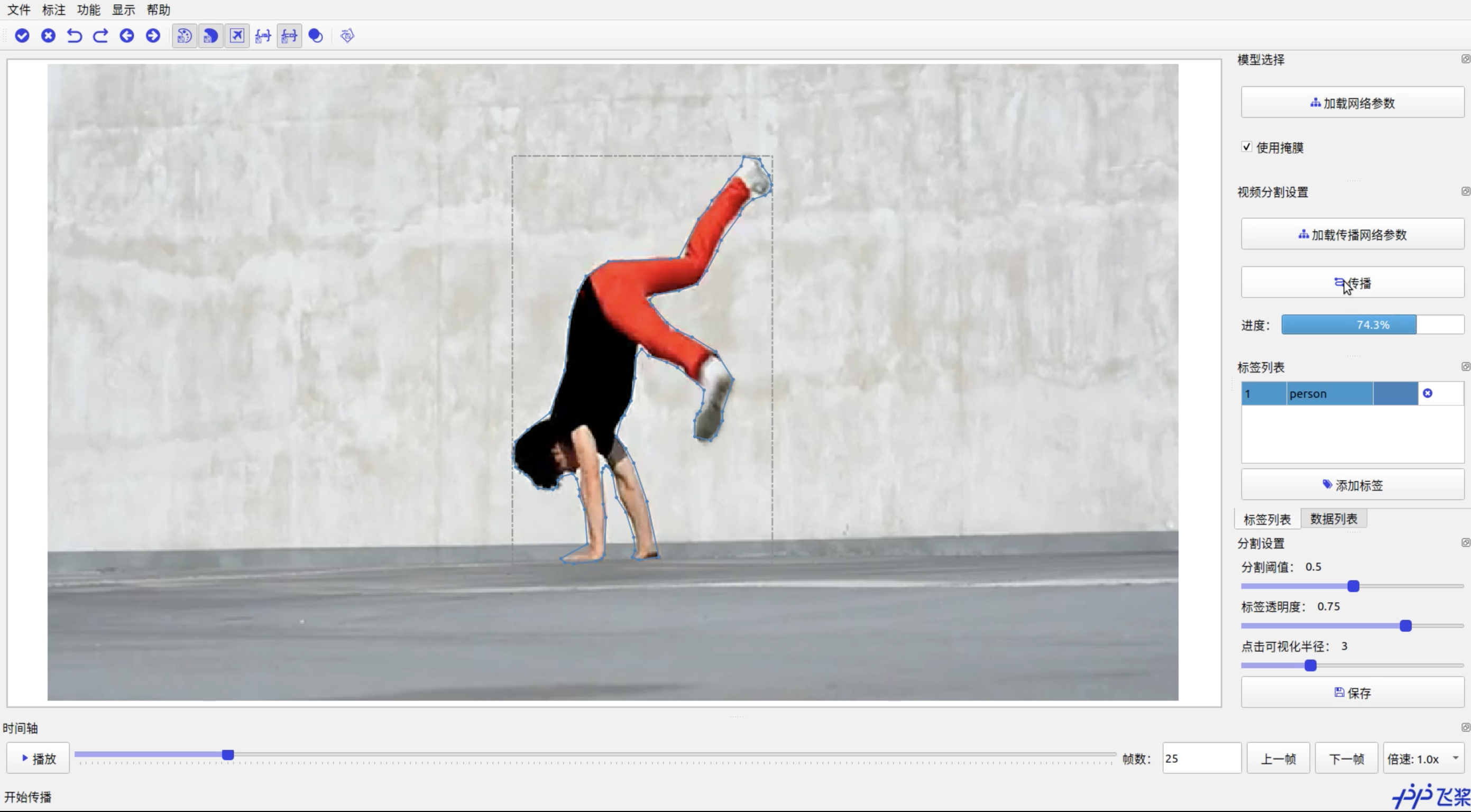}
    \end{overpic}
    \caption{EISeg video annotation UI. After annotating reference frames, click the "Propagate" button to propagate masks to other frames.
    }\label{fig:video}
\end{figure}

\begin{itemize}
\item Loading a model and its parameters. Choose a suitable model listed under the model preparation section in \href{https://github.com/PaddlePaddle/PaddleSeg/tree/release/2.5/EISeg}{EISeg’s documentation} based on the type of images to be annotated. Download and unarchive the zip file. Then put the model structure file ending in pdmodel and the model parameter file ending in pdiparams into the same folder. In the model selection file picker of EISeg, choose the pdiparams file. Static models can take tens of seconds to initialize. Proceed to the next step after the model is loaded. Paths to recently used models will be saved by EISeg to facilitate easy switching. And there’s the option to automatically load the most recent model on software starts.

\begin{table*}[hpt]
	\centering
	
	\caption{NoC metrics of various interactive segmentation algorithms on GrabCut, Berkeley, DAVIS and Pascal VOC dataset. Lower NoC indicates better performance. Best results are bolded.} 
	\label{tab:sota}
	\vspace{0.1in}
	\begin{tabular}{c|c|c|c|c|c|c} 
		\hline
		\multirow{2}{*}{Method} & \multicolumn{2}{c|}{GrabCut~} & Berkeley~     & \multicolumn{2}{c|}{DAVIS~}   & Pascal VOC     \\ 
		\cline{2-7}
		& NoC@85        & NoC@90        & NoC@90        & NoC@85        & NoC@90        & NoC@85         \\ 
		\hline
		GC~\cite{gc}                      & 7.98          & 10.00         & 14.22         & 15.13         & 17.41         & -              \\
		GM~\cite{gm}                      & 13.32         & 14.57         & 15.96         & 18.59         & 19.50         & -              \\
		RW~\cite{rw}                     & 11.36         & 13.77         & 14.02         & 16.71         & 18.31         & -              \\
		ESC~\cite{esc}                     & 7.24          & 9.20          & 12.11         & 15.41         & 17.70         & -              \\
		GSC~\cite{esc}                     & 7.10          & 9.12          & 12.57         & 15.35         & 17.52         & -              \\ 
		\hline
		DOS~\cite{deep1}            & -             & 6.04          & 8.65          & -             & -             & 6.88           \\
		LD~\cite{ld}        & 3.20          & 4.79          & -             & 5.05          & 9.57          & -              \\
		RIS-Net~\cite{ris}                 & -             & 5.00          & 6.03          & -             & -             & 5.12           \\
		ITIS~\cite{itis}                    & -             & 5.60          & -             & -             & -             & 3.80           \\
		CAG~\cite{cag}                     & -             & 3.58          & 5.60          & -             & -             & 3.62           \\
		BRS~\cite{brs}                     & 2.60          & 3.60          & 5.08          & 5.58          & 8.24          & -              \\
		FCA~\cite{fca}                 & -             & 2.08          & 3.92          & -             & 7.57          & 2.69           \\
		IA+SA~\cite{eccv2020}                   & -             & 3.07          & 4.94          & 5.16          & -             & 3.18           \\
		f-BRS-B~\cite{f-brs}                 & 2.50          & 2.98          & 4.34          & 5.39          & 7.81          & -              \\
		RITM-H18~\cite{ritm}                & \textbf{1.54} & \textbf{1.70} & 2.48          & 4.79          & 6.00          & 2.59           \\ 
		\hline
		Our method                & 1.60          & 1.72          & \textbf{2.40} & \textbf{4.54} & \textbf{5.77} & \textbf{2.50}  \\
		\hline
		
	\end{tabular}
	\vspace{-0.15in}
\end{table*}

\begin{figure*}[!t]
	\begin{center}
		\begin{tabular}{ccc}
			
			\includegraphics[width=0.31\linewidth, height=3.7cm]{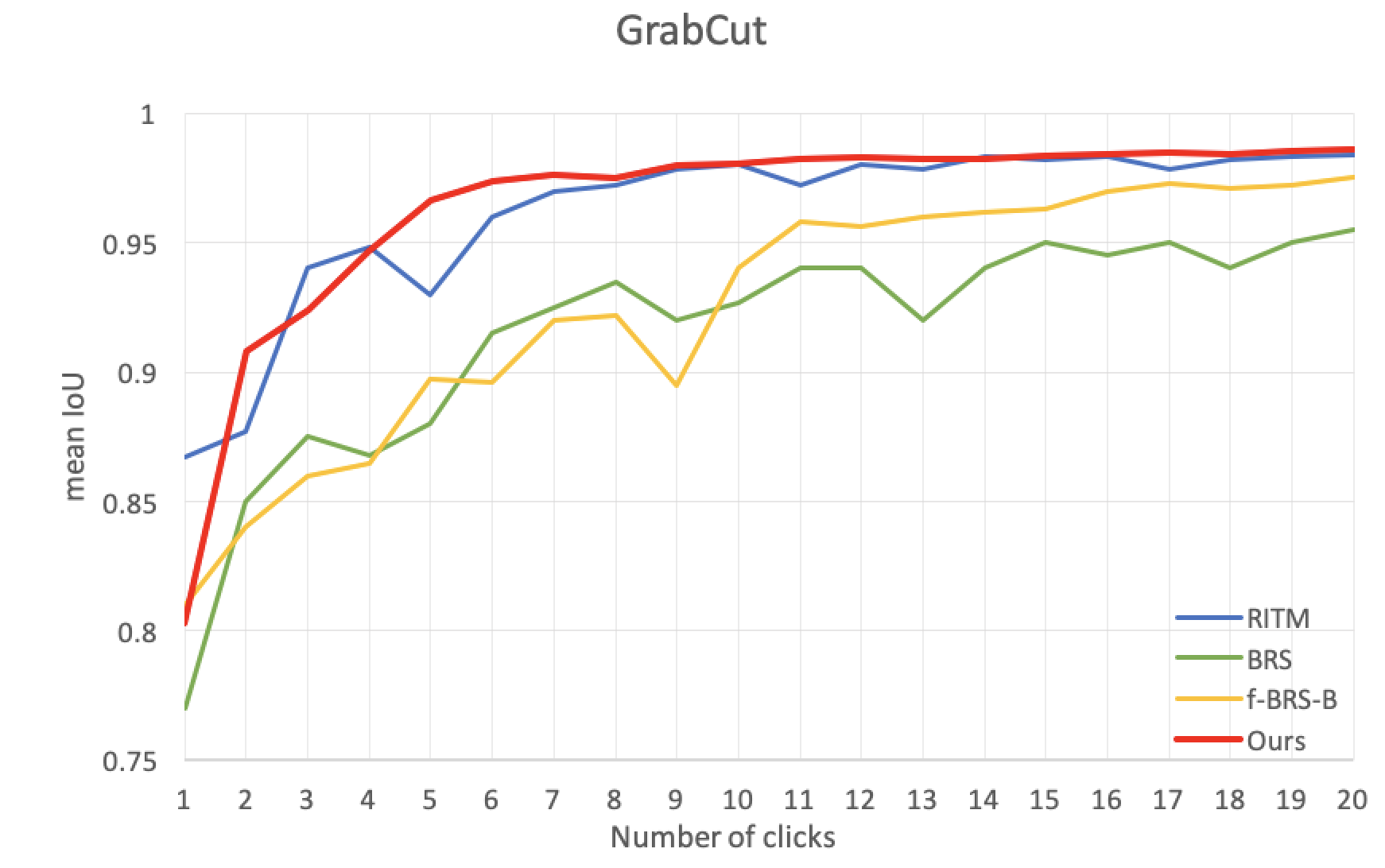}&
			\includegraphics[width=0.31\linewidth, height=3.8cm]{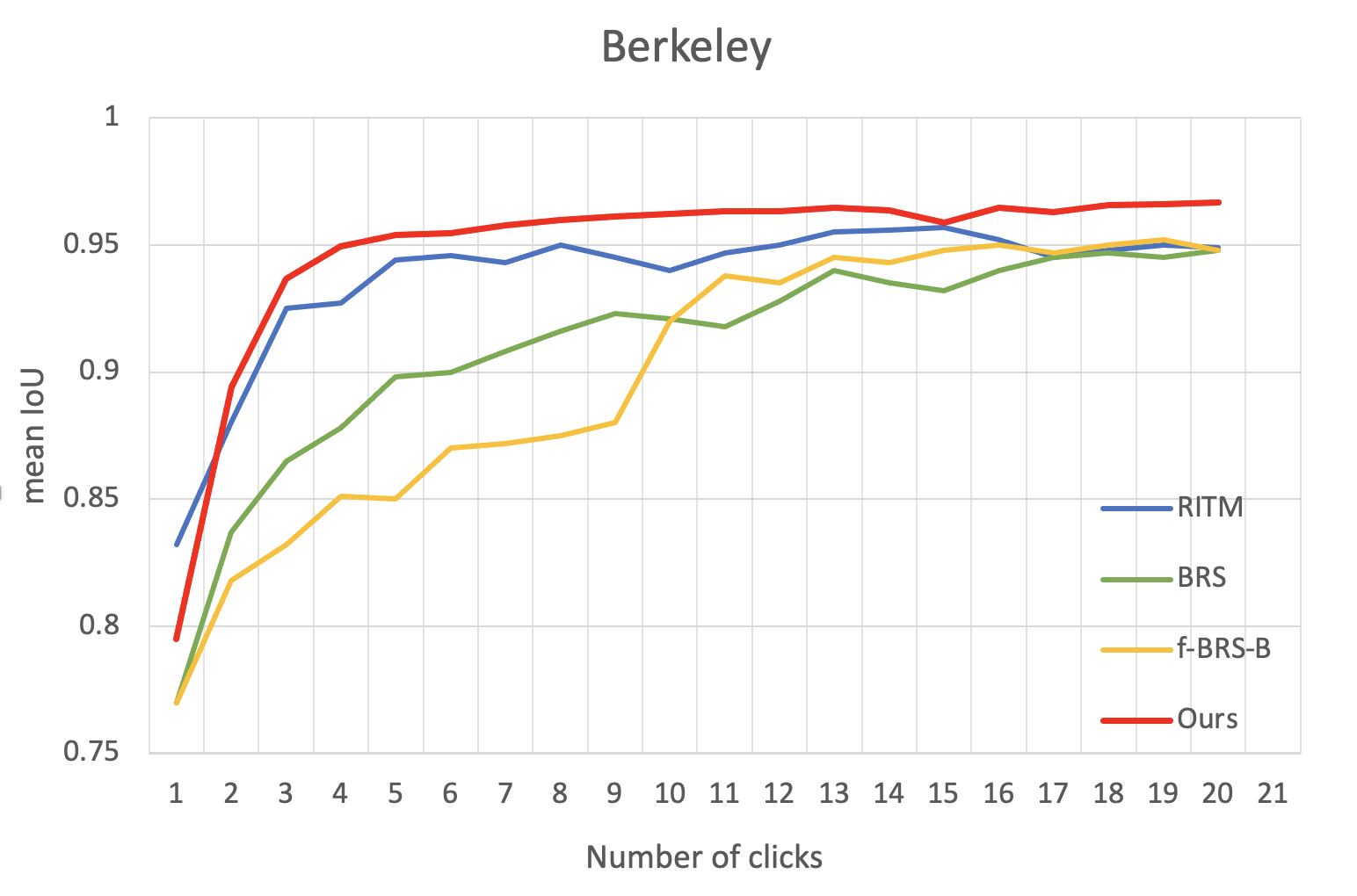}&
			\includegraphics[width=0.31\linewidth, height=3.75cm]{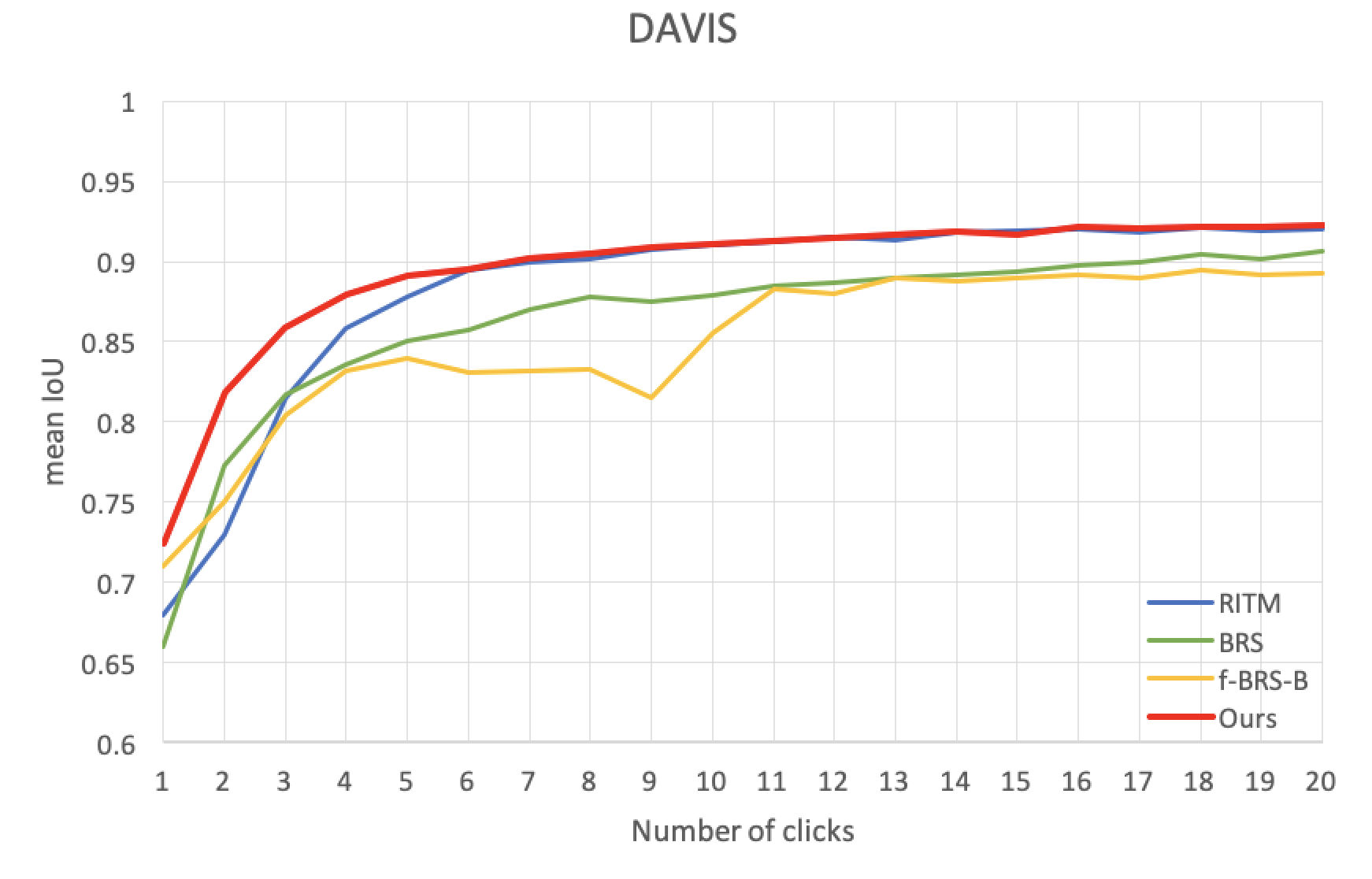}\\
			
		\end{tabular}
	\end{center}
	
    \caption{mIoU curve of various algorithms on GrabCut, Berkeley and DAVIS dataset. First 20 clicks.} 
    \label{figshow}
\end{figure*}

\begin{table*}[hpt]
    \centering
    \caption{Average inference time of EISeg on CPU and GPU.} 
    \label{tab:time}
    \begin{tabular}{c|c|c|c|c|c}
        \hline
        \multirow{2}{*}{Model}                        & \multirow{2}{*}{Model Application Direction} & \multicolumn{2}{c|}{CPU(s)} & \multicolumn{2}{c}{GPU(s)}  \\
        \cline{3-6}
                                                   &                         & First click & Other clicks       & First click & Other clicks       \\
        \hline
        static\_hrnet18\_ocr64\_cocolvis           & Common high precision                   & 0.39      & 0.14           & 0.02    & 0.02              \\
        static\_hrnet18\_ocr64\_cocolvis\_edgeflow & Common high precision                   & 0.61      & 0.50           & 0.04    & 0.03              \\
        static\_hrnet18\_ocr64\_human              & High precision portrait                   & 0.40      & 0.13           & 0.03    & 0.03              \\
        static\_hrnet18\_ocr48\_cocolvis           & Common lightweight                   & 0.20      & 0.09           & 0.02    & 0.01              \\
        static\_hrnet18\_ocr48\_human              & Portrait lightweight                   & 0.18      & 0.09           & 0.02    & 0.01              \\
        static\_hrnet18\_ocr48\_lits               & Liver segmentation model                  & 0.18      & 0.09           & 0.02    & 0.01  \\
        \hline
    \end{tabular}
\end{table*}
\item Opening an image or folder. Choose an image or folder to open. Depending on the type of images opened, EISeg may also show widgets containing domain-specific tools. For remote sensing images in tiff format, a remote sensing toolbox widget containing band settings and geographic metadata viewers will be shown. For medical images in Dicom format, a medical image toolbox widget containing window level and window width adjustment tools will be shown. If the width or height of the opened image exceeds 2000 pixels, a grid annotation widget containing an option to split the image into small patches and annotate them one by one will be shown.

\item Adding/Loading categories. Users can add categories to the project while annotating. The category widget contains four columns, corresponding to the id, comment, color, and deletion of a category. Created categories can be saved to a txt file for other collaborators to load. Project categories will be automatically loaded when opening a folder if the txt file is found under the folder.

\item Annotating. Click the interactive annotation button. While in interactive segmentation mode, a left or right mouse click on the image will add a positive or negative control point, respectively. The region around positive clicks will be considered foreground while the region around negative clicks will be considered background. Users can add multiple clicks until the segmentation mask is satisfactory. Typically only less than 5 clicks will be necessary.  By default, pressing space will finish interactive segmentation for the current object and generate a polygon outline for the segmentation mask. Dragging vertexes of the polygon can change the vertex’s position. Double-clicking a vertex deletes the vertex. Double clicking on the edge of the polygon creates a vertex at the clicked position. 

\item Autosaving. There is the option to automatically save the segmentation mask when switching to the next image. Users just need to enable this feature and configure a folder to save segmentation masks.
\end{itemize}

\subsection{Video and 3D medical image annotation}

EISeg treats the inter-slice direction axis in medical images similar to the temporal axis in videos. A script is provided to users to convert medical images to videos before performing 3D segmentation. The user interface for video segmentation is shown in Figure~\ref{fig:video}. The main steps include:

\begin{itemize}
\item Opening a video or folder. After a video or folder is selected, the first frame of the video will be shown in the canvas and all videos under the folder will be listed in the data list widget. A video annotation toolbox widget containing video playback control and the function to propagate annotation through video will be shown automatically.

\item Annotating reference frames. The process is the same as that for interactively annotating an image described in 4.1\ref{image_annotation}. But note that it's advised to annotate all objects of interest in reference frames for the best propagation result.

\item Loading a propagation model. The process is similar to loading an interactive annotation model. Download a propagation model listed in EISeg documentation matched with the interactive annotation model, unarchive the zip file and choose a model parameter file with a name ending in pdiparams to load.

\item Propagating annotations. Press the frame propagation button in the video toolbox widget. Then the propagation model will generate segmentation masks on frames with objects similar to those in reference frames. If the propagation results aren’t satisfactory, users can annotate more reference frames and perform frame propagation again.

\item Saving. Press the save button in the lower left corner and choose a path to save segmentation results to.

\end{itemize}

\section{Experiments}

\subsection{Experiment Setup}
\textbf{Dataset:} We carried out experiments on four datasets commonly used to evaluate interactive segmentation algorithms's performance. They are GrabCut~\cite{grabcut}, Berke-ley~\cite{berkeley}, Pascal VOC~\cite{pascalvoc}, and DAVIS~\cite{davis}. The GrabCut dataset contains 50 images and 50 corresponding masks. The Berkeley dataset contains 96 images and 100 masks. For Pascal VOC datasets, we only used the validation subset, which contain 1449 images and 3417 instances. Images in the DAVIS dataset are randomly sampled from video object segmentation datasets. We used 345 images and 345 corresponding masks introduced in~\cite{brs,ritm}. All of our experiments are using EISeg 
at PaddleSeg\footnote{https://github.com/PaddlePaddle/PaddleSeg}~\cite{paddleseg} and PaddlePaddle\footnote{https://github.com/PaddlePaddle/Paddle}~\cite{paddlepaddle}.

\textbf{Metrics:} Mean intersection over union (IoU) and the standard number of clicks (NoC) are commonly used metrics for interactive segmentation. NoC represents the number of interactive clicks required for the outputted mask to reach a certain IoU threshold, usually set to 85\% and 90\%~\cite{ritm, dextr, f-brs, brs}, i.e. NoC@85 and NoC@90. During the evaluation phase, we used a simulator to generate clicks similar to~\cite{f-brs, brs, ritm}. The first click will always be a positive one in the object of interest. The next click can be positive or negative, placed at the center of the area with maximum error. The maximum number of clicks is limited to 20. 

\subsection{Comparison with SOTA}

We used NoC@85 and NoC@90 as evaluation metrics. Results on the GrabCut~\cite{grabcut}, Berkeley~\cite{berkeley}, Pascal VOC~\cite{pascalvoc}, and DAVIS~\cite{davis} datasets are shown in Table~\ref{tab:sota}. Our approach achieved the best performance on all datasets except for GrabCut. Images in the GrabCut dataset have a limited number of objectives and clear boundaries, so all the deep-learning-based interactive segmentation algorithms performed very well. Our methods have also produced competitive results on this dataset. Images in the Berkeley dataset have complex boundaries, such as bicycle wheel springs and parachute lines. In this dataset, our method achieved the best performance compared to other methods. Since our method uses edge mask as a priori information, it is able to provide finer details in segmentation results. DAVIS and Pascal VOC datasets contain over 1.5k images from various scenes, which are more suitable for general real-world tasks. Our method achieved the best results on these two datasets. From the experiments, we concluded that the early-late fusion strategy can effectively prevent the fading of user click information as the network gets deeper. Therefore, our method responds effectively to clicks and is robust across scenarios. 

Figure~\ref{figshow} shows the change of mIoU in the first 20 clicks on various datasets. We can conclude from the graph that: 1) Compared with other methods, our method performs the most stable. Our method’s mIoU always increases with the number of clicks. In contrast, other methods have degradation problems. For example, on the GrabCut dataset, RITM's mIoU drops dramatically as the number of clicks increases from 4 to 5 and from 10 to 11. As mentioned above, our proposed edge estimation can improve the stability of segmentation. 2) Compared with other methods, our method requires fewer clicks to achieve a higher mIoU. In the Berkeley dataset, our method’s average NoC@95 is around 4 while that of other methods all exceed 10. 3) After 20 clicks, our method achieves the highest mIoU among all methods. This proves that our method has good segmentation capability.

\subsection{Single Click Speed for Different Models}

We tested the speed of our models on CPU and GPU. The results are shown in Table~\ref{tab:time}. From the table, we can conclude that when inference is run on a GPU, the time it takes is not really noticeable to users. When running on a CPU, aside from the first click, all models are able to produce inference results within 0.5s on average. Users may notice the delay but it won’t be very obvious. The prediction speed of the models allows for running an interactive segmentation application on a CPU.

\section{Summary and Prospect}

In this paper, we present EISeg, an efficient and accurate interactive image segmentation tool. It can rapidly generate highly accurate segmentation masks from reduced user inputs with the help of deep learning in general cases. To assist annotation in more industries,  we also provide domain-specific models for remote sensing, medical imaging, industrial quality inspection, and temporal-aware model for videos. The experiments show that EISeg is fast and also achieves state-of-the-art performance in segmentation accuracy.

{\small

\bibliographystyle{ieee_fullname}

}
\end{document}